\documentclass{article}

\usepackage[preprint]{neurips_2018}

\usepackage[utf8]{inputenc} % allow utf-8 input
\usepackage[T1]{fontenc}    % use 8-bit T1 fonts
\usepackage{hyperref}       % hyperlinks
\hypersetup{
    colorlinks=true,
    linkcolor=blue,
    urlcolor=cyan,
    citecolor=red,
}
\usepackage{url}            % simple URL typesetting
\usepackage{booktabs}       % professional-quality tables
\usepackage{amsfonts}       % blackboard math symbols
\usepackage{nicefrac}       % compact symbols for 1/2, etc.
\usepackage{microtype}      % microtypography

\usepackage{caption} 
\usepackage{algorithm}
\usepackage[noend]{algorithmic}
\usepackage[utf8]{inputenc} % allow utf-8 input
\usepackage[T1]{fontenc}    % use 8-bit T1 fonts

\newcommand{\ltwo}{\ensuremath{L_2}}
\newcommand{\sample}{R}
\newcommand{\asigma}{\tilde{\sigma}} % "acceptable" std dev of noise in the final average
\newcommand{\numvecs}{m}
\newcommand{\numgroups}{G}
\newcommand{\est}[1]{\hat{#1}}  % Final estimate of a vector
\newcommand{\scale}[1]{\mathbf{s}(#1)}
\newcommand{\clip}[2]{\pi_{\!#2}\!\left(#1\right)}

\usepackage{hyperref}
\usepackage{booktabs}       % professional-quality tables
\usepackage{amsfonts}       % blackboard math symbols
\usepackage{times}
\usepackage{subcaption}

\usepackage[shortlabels]{enumitem}
\setlist{nolistsep}
\setlist[itemize]{noitemsep, topsep=0pt}

\usepackage{graphicx} % more modern
\usepackage{xcolor}

\usepackage{array}
\usepackage{amssymb}
\usepackage{amsmath}
\usepackage{amsthm}
\usepackage{xspace}

\newtheorem{definition}{Definition}

\newcolumntype{H}{>{\setbox0=\hbox\bgroup}c<{\egroup}@{}}

\definecolor{darkgreen}{rgb}{0,0.4,0.0}
\definecolor{darkblue}{rgb}{0,0.1,0.3}
\definecolor{darkred}{rgb}{0.7,0.0,0.0}

\definecolor{modelgray}{HTML}{808080}
\definecolor{modelblue}{rgb}{0.2980392156862745, 0.4470588235294118, 0.6901960784313725}
\definecolor{modelred}{rgb}{0.7686274509803922, 0.3058823529411765, 0.3215686274509804}

\newcommand{\noiclr}[1]{}

 \newcommand{\mcm}[1]{}

\newcommand {\norm}[1]{\ensuremath{\| #1 \|}}

\newcommand{\eat}[1]{}

\DeclareMathOperator*{\E}{\mathbb{E}}

\newcommand{\eps}{\epsilon}

\title{A General Approach to Adding Differential Privacy to Iterative Training Procedures}

\author{
  H. Brendan McMahan \\
  \texttt{mcmahan@google.com} \\
  \And
  Galen Andrew \\
  \texttt{galenandrew@google.com}\\
  \And
  \'Ulfar Erlingsson \\
  \texttt{ulfar@google.com}\\
  \And
  Steve Chien \\
  \texttt{schien@google.com}\\
  \And
  Ilya Mironov \\
  \texttt{mironov@google.com}
  \And
  Nicolas Papernot \\
  \texttt{papernot@google.com}
  \And
  Peter Kairouz \\
  \texttt{kairouz@google.com}
}

\begin{document}

\maketitle

\begin{abstract}
In this work we address the practical challenges of training machine learning models on privacy-sensitive datasets by introducing a modular approach that minimizes changes to training algorithms, provides a variety of configuration strategies for the privacy mechanism, and then isolates and simplifies the critical logic that computes the final privacy guarantees. A key challenge is that training algorithms often require estimating many different quantities (vectors) from the same set of examples --- for example, gradients of different layers in a deep learning architecture, as well as metrics and batch normalization parameters. Each of these may have different properties like dimensionality, magnitude, and tolerance to noise. By extending previous work on the Moments Accountant for the subsampled Gaussian mechanism, we can provide privacy for such heterogeneous sets of vectors, while also structuring the approach to minimize software engineering challenges.
\end{abstract}

\section{Introduction}

There has been much work recently on integrating differential privacy (DP) techniques into iterative training procedures like stochastic gradient descent~\citep{chaudhuri11dperm,bassily14focs,abadi16dpdl, WLKCJN17,  PapernotAEGT17}; for completeness we provide a formal definition of DP in Appendix~\ref{sec:dp}. Although these works differ in the granularity of privacy guarantees offered and the method of privacy accounting, most proposed approaches share the general idea of iteratively computing a model update from training data and then applying the Gaussian mechanism for differential privacy to the update before incorporating it into the model. Our goal in this work is to decouple, to the extent possible, three aspects of integrating a privacy mechanism with the training procedure:
\begin{enumerate}[label=\alph*)]
    \item the specification of the training procedure itself (e.g., stochastic gradient descent with batch normalization and simultaneous collection of accuracy metrics and training data statistics), \label{it:user}
    \item the selection and configuration of the privacy mechanisms to apply to each of the aggregates collected (model gradients, batch normalization weight updates, and metrics), and \label{it:mech}
    \item the accounting procedure used to compute a final $(\varepsilon, \delta)$-DP guarantee. \label{it:account}
\end{enumerate}

\begin{NoHyper}
This separation is critical: the person implementing~\ref{it:user} is likely not a DP expert, and this code typically already exists; there are many configuration options for~\ref{it:mech}, which will likely require experimentation, and this configuration logic may become complex; thus isolating the key privacy calculations in~\ref{it:account} and keeping them as simple (and well tested) as possible prevents bugs in~\ref{it:user} or~\ref{it:mech} from introducing errors in the calculation of the actual privacy achieved.
\end{NoHyper}

While model training is our primary motivation, the approach is applicable to any iterative procedure that fits the following template. We have a database with $n$ records. A {\em record} might correspond to a single training example, a ``microbatch'' of examples, or all of the data from a particular user or entity (e.g., to achieve user-level DP as in~\citet{mcmahan2018learning}). On each round, a random subset of records (a {\em sample}) is selected and the training procedure consumes the results of a number of vector queries over that sample; see Table~\ref{tab:definitions}. Such vector queries may include the average gradient for each layer, updates to batch-normalization parameters, or the average value for different training accuracy metrics. We describe a general approach to allocating a privacy budget across each of these queries and analyzing the privacy cost of the complete mechanism, all respecting the decoupling of concerns described earlier. Our analysis builds on the Moments Accountant approach of~\citet{abadi16dpdl}, which applies to a single vector query per round, and generalizes the extension of~\citet{mcmahan2018learning} to multi-vector queries.

\begin{table}[t]
\centering
\begin{tabular}{|l|p{3.5cm}|p{4cm}|p{3.5cm}|}
\hline
             & Per-example SGD & Microbatch SGD & Federated learning \newline (user-level DP) \\ \hline
{\em record} & gradient on one example & average gradient on one \newline {\em microbatch} (\textasciitilde10 examples) & model update from one user \\ \hline
{\em sample} & minibatch (\textasciitilde100 examples) & minibatch (\textasciitilde10 microbatches for a total of 100 examples) & set of participating user devices for the round \\ \hline
\end{tabular}
\caption{Defining {\em record} and {\em sample} in different training contexts.}
\label{tab:definitions}
\end{table}

We focus on the following basic building block for a single vector. Suppose we have a database $X$ with $n$ records consisting of vectors $x^i \in \mathbb{R}^D$ and we are interested in estimating the average\footnote{For simplicity, we focus on unweighted average queries, for example to compute average gradient on a batch of examples; the generalization to weighted average and vector sum queries is straightforward. We also restrict attention to the fixed expected denominator $f_\mathrm{f}$ of \citet{mcmahan2018learning}; extension to other estimators for averages like their $f_\mathrm{c}$ is straightforward.} $\frac{1}{n}\sum_i x^i$. Given a selection probability $q$, clipping threshold $S$, and noise multiplier $z$, the procedure is:
\begin{enumerate}
\item Select a subset of the records $\sample \subseteq [1, \ldots, n]$ by choosing each record with probability $q$. \label{it:sample}
\item Clip each $x^i$ for $i \in R$ to have maximum $\ltwo$ norm $S$
using $\clip{x}{S} = x \cdot \min \left(1, S / \norm{x}_2 \right)$.  \label{it:clip}
\item Output $\hat{x} = \frac{1}{qn} \left( \sum_{i \in \sample} \clip{x^i}{S} + \mathcal{N}(0; \sigma^2 I) \right)$ where $\sigma = zS$.
\end{enumerate}
The quantity $\sum_{i \in \sample} \clip{x^i}{S} + \mathcal{N}(0; \sigma^2 I)$ is the output of the Gaussian mechanism for sums. As $\E[|\sample|] = qn$, scaling it by $1/qn$ produces an unbiased estimate of the average. The {\em noise multiplier} $z \equiv \sigma/S$ (the ratio of the noise to the $\ltwo$-sensitivity of the query) acts as a knob to trade off privacy vs.\ utility. If we choose $z=\frac{1}{\varepsilon}\sqrt{2 \ln 1.25 / \delta}$, the mechanism is $(q\varepsilon, q\delta)$-differentially private with respect to the full database~\citep{BeimelBKN14,dwork14book}. Importantly, the privacy cost of this mechanism is fully specified by $q$ together with the \emph{privacy tuple} $(S, \sigma)$, where $S$ is an upper bound on the $\ltwo$ norm of the vectors being summed, and $\sigma$ is the standard deviation of the noise added to the sum.

We generalize the above procedure to the case where each record corresponds to a collection of vectors. We still do the sampling step \eqref{it:sample} only once, but we estimate the average of each of the vectors separately, potentially with different clipping thresholds and noise multipliers. Let $v = (v_1, \dots, v_\numvecs)$ be the total set of vectors for which averages are to be estimated privately. (Note, we will include the superscript $i$ on the $i$\textsuperscript{th} record $v^i$ only when necessary in equations that sum over records.) In general, we may partition this set of $\numvecs$ vectors into multiple groups, e.g., fully connected layers vs.\ convolutional layers vs.\ metrics. We assume the user (that is, the person using the privacy tools defined here) has identified the relevant set of groups whose averages are needed in the training procedure. For each of these, they need to specify a privacy mechanism together with some hyperparameters. We first describe the privacy mechanisms that can be applied to individual vectors or groups of vectors, then show how the privacy cost of the full collection of mechanisms can be calculated, and finally propose strategies for choosing the parameters to achieve the desired privacy versus utility tradeoff.

Implementations of techniques in this paper
may be found in 
the open-source TensorFlow Privacy framework~\citep{tfprivacy}
for TensorFlow~\citep{tensorflow2015},
as described in Section~\ref{sec:tfp}.

\section{Privacy mechanisms for a group of vectors}
In this section, we describe two strategies that can be applied to a single group of vectors, {\em WLOG} the first $k$, $(v_1, \dots, v_k)$, for $k \le \numvecs$; when $k=1$, the two mechanisms described are identical. Both mechanisms allow individual noise standard deviation parameters to be used for the separate groups. While this might at first seem to preclude the use of the Moments Accountant, which requires spherical noise, we will show how to resolve this issue in the next section.\footnote{Privacy mechanisms for groups can be used within TensorFlow Privacy~\citep{tfprivacy} by employing the \texttt{NestedQuery} class, which evaluates an arbitrary nested structure of queries where each leaf query would be a \texttt{GaussianAverageQuery} corresponding to one group of vectors.}

\paragraph{Separate clipping and noise parameters.}
This strategy essentially treats the whole group as a single concatenated vector $v = \left(v_1, \ldots, v_k\right)$. The user provides $S_g$, a clipping parameter, and $\sigma_g$, a noise parameter. For now, assume both of these parameters are simply chosen so as to provide reasonable utility for the resulting average; we will discuss strategies for choosing these parameters in detail in Section~\ref{sec:choosingparams}. The output of the mechanism is
\begin{equation*}
\est{v}_j 
  = \frac{1}{qn} \sum_{i \in \sample} \clip{v^i}{S_g}_j 
      + \mathcal{N}(0; \sigma_g^2 I)
  = \frac{1}{qn} \left( \sum_{i \in \sample} \clip{v^i}{S_g}
      + \mathcal{N}(0; \asigma_g^2 I)\right)_j,    
\end{equation*}
where $\asigma_g=qn \sigma_g$ and $j\in[k]$. The final expression shows that the mechanism is equivalent to the Gaussian mechanism for sums with privacy tuple $(S_g, \asigma_g)$. Applying this mechanism with $S_g = S^*$ to all $\numvecs$ vectors recovers the ``flat clipping'' approach of \citet{mcmahan2018learning}, and applying this mechanism separately to each of the vectors with $S_g = S^*/\sqrt{\numvecs}$ recovers their ``per-layer clipping'' approach (where $S^*$ is the total $\ltwo$ bound). Another reasonable strategy that takes into account dimensionality is to apply the mechanism separately with $S_g = S^*/\sqrt{d_g/D}$ where $d_g$ is the dimensionality of $v_g$.

\newcommand{\scp}{\alpha}
\paragraph{Joint clipping.}
Here we introduce a new mechanism that allows us to clip less aggressively than applying the previous strategy to each vector individually, while still letting different vectors live on different multiplicative scales. The user supplies as input scale parameters $\scp_1, \dots, \scp_k$, which may be thought of as bounds or reasonable $\ltwo$ norm clip parameters on the individual $v_i$, were they to be clipped individually. The strategy first does a pre-processing step via the scaling operator $\scale{v; \scp_{1:k}} = (v_1/\scp_1, \dots, v_k/\scp_k)$. If $\norm{v_j}_2 \le \scp_j$ for all $j\in[k]$, then the joint norm $\norm{\scale{v;\scp_{1:k}}}_2 \le \sqrt{k}$, however it may typically be much less. Then joint clipping and noising is performed using a total clipping parameter $S_g \in \big[0, \sqrt{k}\big]$ and noise with the standard deviation of~$\alpha_j \sigma_g$. The mechanism's output then scales the vectors back by the $\scp_j$ factor in post-processing:
\begin{align*}
\est{v}_j 
  &= \frac{1}{qn} \sum_{i \in \sample} \scp_j\clip{\scale{v^i;\scp_{1:k}}}{S_g}_j 
       + \mathcal{N}\left(0; (\scp_j \sigma_g)^2 I \right)\\  %\label{eq:jointest}
  &= \frac{\scp_j}{qn} \left( \sum_{i \in \sample} \clip{\scale{v^i;\scp_{1:k}}}{S_g}
       + \mathcal{N}\left(0; \asigma_g^2 I\right)\right)_j,   %\label{eq:jointpriv}
\end{align*}
where again $\asigma_g=q n \sigma_g.$ The final expression shows the output can be written as a post-processing of the subsampled Gaussian mechanism for sums with privacy tuple $(S_g, \asigma_g)$. Note that if no clipping happens then $\scp_j\clip{\scale{v; \scp_{1:k}}}{S_g}_j  = v_j$ for all $j$.

To see where this mechanism might be superior to the first, suppose $v_1$ and $v_2$ have $\norm{v_1}_2 \le 1$ and $\norm{v_2}_2 \le 100$, and suppose they can tolerate noise standard deviations of $0.01$ and $1$ respectively. Additionally, assume it is known that either $v_1^i$ or $v_2^i$ will be zero for any record $i$. We could clip these separately, but this ignores the (useful) side information that one of the vectors is always zero. On the other hand, if we treat them as a single group, we cannot take into the account the fact they are on very different scales; in particular, we must pick a single noise value which will either be insufficient to add privacy for $v_2$, or will completely obscure the signal in $v_1$. The joint mechanism proposed here lets us directly handle this situation using $\alpha_1=1$, $\alpha_2=100$, $\sigma_g = 0.01$, and $S_g=1$.

\newcommand{\SG}{S^*}
\section{Composing privacy guarantees for multiple vector groups}\label{sec:composing}
Now, suppose we have partitioned the $\numvecs$ vectors into $\numgroups$ groups, and selected a privacy mechanism for each one, producing privacy tuples $(S_g, \asigma_g)$ for $g \in \{1, \dots, G\}$. From a privacy accounting point of view, each of these mechanisms is equivalent to running a Gaussian sum query on vectors $w_g$ with $\norm{w_g} \le S_g$ and then adding noise $\asigma_g$ to the final sum. We now demonstrate a transformation that lets us analyze this composite mechanism as a single Gaussian sum query on the sample for use with the privacy accountant.
\mcm{I think we likely should re-work this later to work for arbitrary queries of sensitivity $S_g$, but out of scope for now.}

First, we scale each vector $\scale{w;\asigma_{1:G}} = (\frac{w_1}{\asigma_1}, \dots, \frac{w_G}{\asigma_G})$, so 
$\norm{\scale{w;\asigma_{1:G}}} \le 
\SG \equiv \sqrt{\sum_g \left(S_g / \asigma_g\right)^2}.$ 
Now, we imagine a single Gaussian sum query with noise standard deviation $\sigma = 1$, and output the estimate after rescaling by the $\asigma_g$ factors. This is equivalent since
\begin{equation}
\hat{w}_g 
 = \frac{1}{qn} \left( \sum_{i \in \sample} w^i_g 
    + \mathcal{N}(0; \asigma_g^2 I)\right) 
 = \frac{\asigma_g}{qn} \left( \sum_{i \in \sample} \scale{w^i;\asigma_{1:G}} 
     + \mathcal{N}(0; I) \right)_g. \label{eq:equivalence}
\end{equation}
The final expression is a simple post-processing on the output of a single Gaussian sum query with parameters $(\SG, \sigma=1)$. Thus, we can apply the privacy accountant to bound the privacy loss of iterative applications of this mechanism. 

\section{Hyperparameter selection strategies}\label{sec:choosingparams}

Here we consider selecting hyperparameters $q$, $S_g$, and $\sigma_g$ to achieve a particular privacy vs.\ utility tradeoff. Recall for both mechanisms, $\asigma_g = q n \sigma_g$, so the key quantity is 
\[
z = \frac{1}{\SG}
  = \left(\sum_g \left(S_g / \asigma_g \right)^2 \right)^{-1/2}
  = q n\left(\sum_g \left(S_g/\sigma_g \right)^2 \right)^{-1/2}.
\]
Typically, a value of $z \approx 1$ will provide a reasonable privacy guarantee. If $z$ is too small for the desired level of privacy, the user has several knobs available: clip more aggressively by decreasing the $S_g$'s; noise more aggressively by scaling up the $\sigma_g$'s; or increasing $q$. When datasets are large and the additional computational cost of processing larger samples $\sample$ is affordable, this last approach is generally preferable, as observed by \citet{mcmahan2018learning}. If additionally the total number of iterations $T$ is known, then since the privacy cost scales monotonically with any of these adjustments to $z$, a binary search can be performed using the privacy accountant repeatedly with different parameters to find e.g. the precise value of $q$ needed to achieve a particular $(\varepsilon, \delta)$-DP guarantee.

\paragraph{Choosing $\sigma_g$ and $S_g$.}
Typical approaches to setting $S_g$ include: 1) using an \textit{a priori} upper bound on the $\ltwo$ norm; 2) choosing $S_g$ so that ``few'' vectors are clipped; or 3) running parameter tuning grids to find a value of $S_g$ that does not reduce utility (e.g., the accuracy of the model) by too much. If private data is used in 2) or 3), the privacy cost of this should be accounted for. \mcm{Cite some papers on private hyperparameter tuning.} Similar strategies can be used to choose $\sigma_g$, e.g., selecting a value that will introduce an \textit{a priori} acceptable amount of error, or more likely for model training, running experiments to find the largest amount of noise that does not slow the training procedure.

In some cases one may have bounds $S_g$ on the norms of $G$ groups plus an overall target value of $z$, which needs to be distributed across multiple groups.
To achieve {\em proportional} noise, where $\asigma_g \propto S_g$ for all $g$, we can use $\asigma_g = z \sqrt{G} S_g$. Another reasonable alternative, {\em dimensionality adjusted} noise assigns noise proportional to the maximum root mean squared value of the components of $w_g$ given its bound and its dimensionality: $\asigma_g = z \sqrt{D/d_g} S_g$, where $d_g$ is the dimensionality of group $g$ and $D \equiv \sum_g d_g$.

\section{Sampling policies}

The basic update step of the SGD algorithm operates on a small subset of records (the \emph{minibatch}). Convergence guarantees of the standard optimization theory hold under the assumption that each minibatch is an i.i.d.~sample of the training dataset, and the original Moments Accountant by~\citet{abadi16dpdl} supported privacy analysis in this regime.

In practice, there are valid reasons for using alternative policies for sampling minibatches, with implications for privacy analysis. We list three of the most common sampling policies below.

\paragraph{Minibatches are i.i.d.~samples.} Privacy of this sampling procedure is analyzed by~\citet{abadi16dpdl} and it is used by the federated learning framework where decisions of whether to participate in a particular update step are made locally~\citep{mcmahan2018learning}. If the privacy accountant is dependent on the secrecy of the sample (as in the case of the Moments Accountant), then the size of the sample cannot be released without applying a privacy-preserving mechanism, which can be as simple as additive noise. The variability of the sample's size makes this sampling policy a poor fit for hardware accelerators. It can be repaired by sampling subsets of a fixed size from the training set \emph{without replacement}, which leads us to the next policy.

\paragraph{Minibatches are equally sized and independent.} The basic SGD corresponds to this sampling policy and minibatches of cardinality 1. Recent works analyze composition of this sampling policy with a mechanism satisfying RDP~\citep{WBK18} or tCDP~\citep{BDRS18}. Independence of minibatches makes analysis of multiple iterations of SGD straightforward via application of composition rules for differential privacy.

\paragraph{Minibatches are equally sized and disjoint.} In practice, the most common manner of forming minibatches is permuting the training dataset and partitioning it into disjoint subsets of a fixed size. After a single pass (an \emph{epoch}) the process is repeated. This sampling policy can be efficiently implemented, and has intuitive semantics: an epoch corresponds to a training cycle when all examples were visited exactly once. Quantitatively tight analysis of DP-SGD in this model is not known. (A related problem of analyzing \emph{randomized response} followed by a random permutation is addressed by~\citet{shuffling19}.)  

\section{Privacy ledger}
In principle, privacy accounting (via, e.g.\ the moments accountant) could be done in tandem with calls to the mechanism to keep an online estimate of the $(\eps, \delta)$ privacy guarantee. However we advocate a different approach which cleanly separates concerns \ref{it:mech} and \ref{it:account} from the introduction. We maintain a {\em privacy ledger} and record two types of events: {\em sampling events}, which record that a set $R$ of records has been drawn using parameters $q$ and $n$, and {\em sum query} events, which record that a Gaussian sum query has been performed over some group of vectors with privacy tuple $(S_g, \asigma_g)$. Then the privacy accountant can process the ledger \textit{post hoc} to produce a privacy guarantee, first converting each group of one sampling event plus some sum query events to an equivalent single sum query event with parameters $(S^*, \sigma=1)$ using Equation~\eqref{eq:equivalence}.

There are two main advantages of this approach. First, bugs in the hyperparameter selection strategy code cannot affect the privacy estimate. Second, it allows the privacy accounting mechanism to be changed and the ledger reprocessed if, for example, a tighter bound on the privacy loss is discovered after the data has been processed.

\section{TensorFlow Privacy}\label{sec:tfp}

TensorFlow Privacy~\citep{tfprivacy}\footnote{Available from \url{https://github.com/tensorflow/privacy} under Apache 2.0 license.} is a Python library that implements TensorFlow optimizers for training machine learning models with differential privacy. The library comes with tutorials and analysis tools for computing the privacy guarantees provided.
From an engineering perspective, the implementation of differentially private optimizers found in the library leverages the decoupled structure outlined above to make it  easier for developers to both (a) wrap most optimizers into their differentially private counterpart and (b) compare different privacy mechanisms and accounting procedures. 

Perhaps the library is best illustrated by one of its main use cases: training a neural network with differentially-private stochastic gradient descent~\citep{abadi16dpdl}. Given the  stochastic gradient descent optimizer class, \texttt{tf.train.GradientDescentOptimizer}, implemented in the main TensorFlow library, one first wraps it into a new optimizer that implements logic for both the clipping and noising of gradients needed to obtain privacy. This is done by having the optimizer estimate the gradients via an instance of a class implementing the \texttt{DPQuery} interface. A \texttt{DPQuery} is responsible for clipping gradients computed by the optimizer, accumulating them, and returning their noisy average to the optimizer. This introduces two additional hyperparameters to the optimizer: the \textit{clipping norm} and the \textit{noise multiplier}. The \texttt{PrivacyLedger} class  maintains a record of the sum query events for each sampling event which can then be processed by the RDP accountant.

In addition, our implementation leverages microbatches, as defined in Table~\ref{tab:definitions}. This implies that gradients are computed over several examples before they are clipped, and once all microbatches in a minibatch have been processed, they are averaged and noised. This introduces a third additional hyperparameter to the optimizer: the \textit{number of microbatches}. Increasing it often  improves  utility but typically slows down training. 

TensorFlow Privacy is also designed to work with training in a federated context in the vein of~\citet{mcmahan2018learning}. In that case the ``gradients'' supplied to the \texttt{DPQuery} would in fact be the model updates supplied by the users in a given round.

Finally, to compute the $(\varepsilon,\delta)$ differential privacy guarantee for the model, an implementation of the RDP accountant is provided. Given the sampling fraction $q$ and the noise multiplier $z$, the RDP is computed for a step. Summing the RDP over the steps, it can then estimate $\varepsilon$ for a fixed $\delta$.

\section{Floating-point arithmetic and randomness source}

The hallmark feature of the definition of differential privacy is that it is \emph{uncoditional}, in other words, it makes no assumptions about the adversarial knowledge or capabilities. It also puts a high burden on a differentially private implementation: its output distribution must have effectively infinite entropy. In practice, the distribution is defined over only a finite domain (such as a vector of single-precision floating-point numbers) and the source of randomness is guaranteed (at best) to be computationally secure. We consider these issues in turn.

\paragraph{Floating-point arithmetic.} The problem of achieving differential privacy by means of standard floating-point arithmetic has been addressed for the additive Laplace mechanism by~\citet{mironov2012significance}. We leave open the task of developing a provable floating-point implementation of DP-SGD and integrating it into an ML library.

\paragraph{Sources of randomness.} Most computational devices have access only to few sources of entropy and they tend to be very low rate (hardware interrupts, on-board sensors). It is standard---and theoretically well justified---to use the entropy to seed a cryptographically secure pseudo-random number generator (PRNG) and use the PRNG's output as needed. Robust and efficient PRNGs based on standard cryptographic primitives exist that have output rate of gigabytes per second on modern CPUs and require a seed as short as 128 bits~\citep{salmon2011parallel}. 

The output distribution of a randomized algorithm~$A$ with access to a PRNG is indistinguishable from the output distribution of $A$ with access to a true source of entropy \emph{as long as the distinguisher is computationally bounded}. Compare it with the guarantee of differential privacy which holds against any adversary, no matter how powerful. As such, virtually all implementations of differential privacy satisfy only (variants of) Computational Differential Privacy introduced by~\citep{Mironov-CDP}. On the positive side, a computationally-bounded adversary cannot tell the difference, which allows us to avoid being overly pedantic about this point.

A training procedure may have multiple sources of non-determinism (e.g., dropout layers or an input of a generative model) but only those that are reflected in the privacy ledger must come from a cryptographically secure PRNG. In particular, the minibatch sampling procedure and the additive Gaussian noise must be drawn from a PRNG for the trained model to satisfy computational differential privacy. In contrast, microbatches need not be chosen using a randomized process.

\section{Conclusion}
We have shown how the Gaussian mechanism can be applied to vectors of different types with different norm bounds and noise standard deviations, enabling training over heterogeneous parameter vectors, as well as simultaneous privacy-preserving estimation of other statistics such as classifier accuracy, or the number of instances in each class. By implementing iterative training algorithms in terms of a series of Gaussian sum queries and then recording for each query privacy events to a ledger to be processed by a privacy accountant, we separate the three major concerns of implementing privacy-preserving iterative training procedures while allowing flexibility in the specification of clipping strategy and noise allocation. The techniques described in the paper can be easily implemented using the Tensorflow Privacy library.

\begin{small}
\bibliography{dp_gauss.bib}
\end{small}

\appendix
\section{Differential Privacy} \label{sec:dp}

The formal definition of $(\varepsilon, \delta)$-differential privacy is provided here for reference:

\begin{definition}
A randomized mechanism $\mathcal{M} \colon \mathcal{D} \mapsto \mathcal{R}$ satisfies {\bf $(\varepsilon, \delta)$-differential privacy} if for any two adjacent datasets $X, X' \in \mathcal{D}$ and for any measurable subset of outputs $\mathcal{Y} \subseteq \mathcal{R}$ it holds that $\Pr\left[\mathcal{M}(X) \in \mathcal{Y}\right] \leq e^\varepsilon \Pr\left[\mathcal{M}(X') \in \mathcal{Y}\right] + \delta.$
\end{definition}

The interpretation of {\em adjacent datasets} above determines the unit of information that is protected by the algorithm: a differentially private mechanism guarantees that two datasets differing only by addition or removal of a single unit produce outputs that are nearly indistinguishable. For machine learning applications the two most common cases are {\em example-level} privacy (e.g.,~\citet{chaudhuri11dperm, bassily14focs, abadi16dpdl, WLKCJN17, PapernotAEGT17}), in which an adversary cannot tell with high confidence from the learned model parameters whether a given example was present in the training set, or {\em user-level} privacy (e.g.,~\citet{mcmahan2018learning}) in which adding or removing an entire user's data from the training set should not substantially impact the learned model. It is also possible to consider $X$ and $X'$ to be adjacent if they differ by {\em replacing} a training example (or an entire user's data) with another, which would increase the $\varepsilon$ by a factor of two.

\end{document}